\documentclass[10pt,conference,letterpaper]{IEEEtran}
\IEEEoverridecommandlockouts

\usepackage[letterpaper, text={6.5in,9in}, centering]{geometry}
\usepackage{amsmath,amsfonts}    %
\usepackage{textcomp}            %
\usepackage{xcolor}              %
\usepackage{algorithm}           %
\usepackage{algpseudocode}       %
\usepackage{subcaption}          %
\usepackage{graphicx,wrapfig}
\usepackage{multirow}
\usepackage{enumitem}
\usepackage{tabularx}
\usepackage{booktabs}
\usepackage{xspace}
\usepackage{ulem}
\usepackage{tikz}
\usepackage{array}
\usepackage{stfloats}
\usepackage{balance}
\usepackage{cite}                %

\usetikzlibrary{arrows.meta,positioning,calc,fit}
\usepackage{overpic}
\renewcommand{\baselinestretch}{1.0}
\AtBeginDocument{\fontsize{10}{11}\selectfont}

\setlist{nosep,leftmargin=*,itemsep=1pt,topsep=2pt,parsep=0pt,partopsep=0pt}

\usepackage[hidelinks]{hyperref}
\usepackage[capitalize,noabbrev,nameinlink]{cleveref}

\crefformat{section}{\S#2#1#3} 
\crefformat{subsection}{\S#2#1#3}
\crefformat{subsubsection}{\S#2#1#3}
\crefformat{figure}{Fig.~#2#1#3}
\crefformat{equation}{Eqn.~#2#1#3}

\usepackage{siunitx}
\usepackage[most]{tcolorbox}
\tcbset{
  colback=gray!5,
  colframe=gray!60,
  boxrule=0.3pt,
  arc=1.2mm,
  left=6pt,right=6pt,top=6pt,bottom=6pt
}

\newcolumntype{C}{>{\centering\arraybackslash}X}

\renewcommand{\arraystretch}{0.95}

\usepackage{titlesec}
\titlespacing*{\section}{0pt}{2ex plus 0.5ex minus 0.2ex}{1ex plus 0.2ex}
\titlespacing*{\subsection}{0pt}{1.5ex plus 0.4ex minus 0.2ex}{0.6ex plus 0.2ex}
\titlespacing*{\subsubsection}{0pt}{1ex plus 0.3ex minus 0.1ex}{0.4ex plus 0.1ex}

\usepackage{enumitem}
\setlist{nosep,leftmargin=1.5em,topsep=2pt}

\newcommand{\ad}[1]{%
  {\color{orange}%
    \tcbox[on line, colback=white, colframe=orange, boxrule=0.8pt, arc=2pt, boxsep=1pt, left=1pt, right=1pt, top=1pt, bottom=1pt]{%
      \textbf{\color{orange}Alex:}%
    }%
    #1%
  }%
}

\newcommand{\KR}[1]{%
  {\color{purple}%
    \tcbox[on line, colback=white, colframe=orange, boxrule=0.8pt, arc=2pt, boxsep=1pt, left=1pt, right=1pt, top=1pt, bottom=1pt]{%
      \textbf{\color{purple}Kishore:}%
    }%
    #1%
  }%
}

\newcommand\jordan[1]{}

\renewcommand{\ad}[1]{}
\renewcommand{\KR}[1]{}

\begin{document}

\title{Scalable Edge-assisted Fusion and Path Prediction for Connected Autonomous Vehicles}

\author{
\IEEEauthorblockN{Tyler Landle\IEEEauthorrefmark{1}, Jackson Isenberg\IEEEauthorrefmark{1}, Abhijit Chatterjee\IEEEauthorrefmark{1}, Alexandros Daglis\IEEEauthorrefmark{2}\IEEEauthorrefmark{1}, Umakishore Ramachandran\IEEEauthorrefmark{1}}
\IEEEauthorblockA{\IEEEauthorrefmark{1}Georgia Institute of Technology \quad \IEEEauthorrefmark{2}University of Edinburgh \\ \{tlandle3, jisenberg3, adaglis, rama\}@gatech.edu, abhijit.chatterjee@ece.gatech.edu}
\thanks{Extended version of the paper to appear at the ACM/IEEE Symposium on Edge Computing (SEC), 2026.}
}

\maketitle
\thispagestyle{plain}
\pagestyle{plain}

\begin{abstract}
The planning algorithms inside an Autonomous Vehicle (AV) rely on information from on-board sensors whose %
line of sight is limited by emerging traffic conditions and occlusions.
Edge-assisted creation of a unified world model \textit{fusing} information from AVs and Road Side Units (RSUs) in a geographical \textit{locale}, and the prediction of AVs' future trajectories %
can enhance the planning algorithms inside AVs to 
improve quality metrics, such as better traffic flow and collision prevention. AVs participating in such enhancements are called Connected Autonomous Vehicles (CAVs). 
However, such information generated by the edge (world model and motion predictions) must reach the planners within a tight
\textit{Age of Information (AoI)} time budget to be useful.
The state of the art fuses per-CAV information: each AV fuses inputs from other actors locally, which limits both scalability with actor count and quality of results.

We present \textit{Conductor}, an edge-based solution for creating a \textit{unified world model} from the perspective of a fixed 
anchor (e.g., an RSU) in a \textit{locale} and predicting future trajectories of AVs in that \textit{locale}.  Our solution adheres to the AoI time budget by dynamically limiting the number of AVs that would lead to the best quality of results.  Specifically, we introduce an occlusion-aware selector that favors information contribution by AVs that detect objects in the \textit{locale} not covered by RSUs. We pair this selector with a runtime controller that adapts both the number of AV inputs to fuse and the amount of trajectory predictions in each cycle to stay within the AoI time budget.
Evaluation on %
CAV simulation infrastructure shows our joint selector-controller meets the AoI safety bound across traffic scenarios with up to 31 CAVs, with fusion fidelity close to an Oracle and much better than a random selector under the same AoI constraint.

\end{abstract}

\section{Introduction}
\label{sec:intro}

Autonomous vehicles rely on their onboard sensors to plan safely. These sensors have limited range and line of sight. Buildings, trucks, parked vehicles, and traffic infrastructure can hide moving actors until the remaining braking margin is small. Connected Autonomous Vehicles (CAVs) and Roadside Units (RSUs) can help by sharing what they observe. An edge server near the intersection can combine these observations into a shared world model and predict how nearby actors will move.

This assistance is useful only if the shared model's \emph{Age of Information} (AoI) when it reaches each AV is still sufficiently fresh. %
AoI is dictated by the aggregate time spent on sensor acquisition, upload to the edge, edge processing, downlink, deserialization, and waiting for the CAV's next planner cycle to use the information. 
For urban driving, a sub-300 ms AoI---i.e., 4\,m of travel or one lane width, at 14 m/s---allows for a safe braking margin. We therefore use 300\,ms as the AoI upper bound and validate its effect in closed-loop driving experiments. %

Existing cooperative perception systems make different tradeoffs. Early fusion sends raw sensor data to the edge, preserving information but requiring prohibitive uplink bandwidth at dense intersections. Late fusion sends compact object detections, but loses spatial details that could help recover occluded objects. Feature-level (intermediate) fusion strikes a middle ground: each CAV sends compressed features rather than raw point clouds or final object lists, preserving spatial information at a manageable payload size.

Leveraging the strengths of intermediate fusion, we study a new architecture that employs an edge component to build \textit{one} shared world model for a geographic \textit{locale}, anchored at a fixed RSU. The model is valid for all the CAVs in that \textit{locale}, so the edge fuses AV-contributed observations once and sends the same update to all recipients. This approach obviates duplicate per-AV fusion and enables one multicast update per cycle. It also creates a shared timing dependency: if the edge update is late, all recipients that depend on that update consume older shared state or fall back to onboard perception. Freshness is therefore a system-level property, not only a per-vehicle latency metric.

\begin{figure}[t]
  \centering
  \includegraphics[width=\columnwidth]{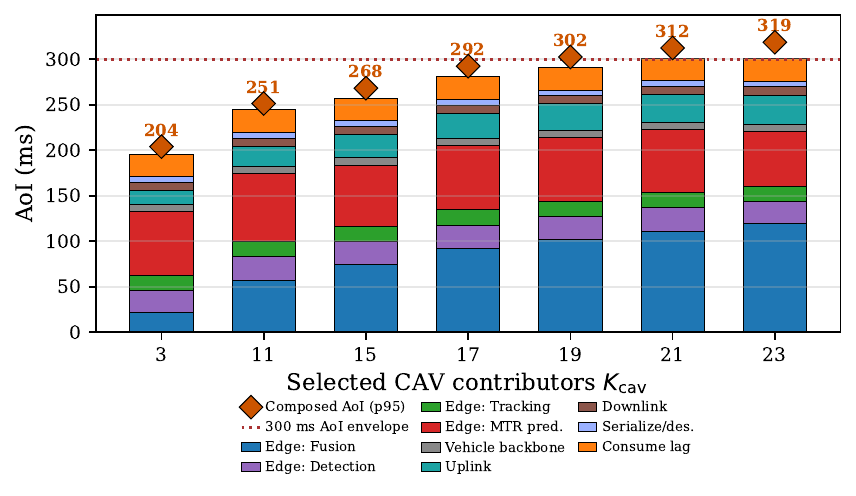}
  \vspace{-6mm}
  \caption{Deadline cliff as more AVs participate in fusion. The baseline fuses every available AV on each update and runs on edge-class hardware. Fusion grows with the number of participating AVs and becomes the largest edge cost beyond four AVs. Stacked bars show mean per-component latency contribution and diamond markers show p95 AoI, %
  crossing the 300\,ms bound between 17 AVs (292\,ms) and 19 AVs (302\,ms).}
  \label{fig:cliff_per_stage} 
\end{figure}

This design raises the central question studied in the paper: once fusion, tracking, and trajectory prediction run together at the edge, can the service keep the world model fresh as the number of participating AVs grows? 
As shown in \cref{fig:cliff_per_stage}, p95 AoI (the sum of sensor upload time, edge processing time, time to disseminate to the CAVs, and the wait time until the vehicle planner runs next) crosses the 300\,ms safety bound beyond 17 CAVs.
We call this sharp loss of AoI compliance the \textit{deadline cliff}.

We present \emph{Conductor}, an edge service designed to keep the shared world model construction within the AoI budget. 
Conductor's key insight is that not all AVs are equally useful in constructing the world model. The RSU already sees much of the intersection, so many AVs duplicate its view while still increasing fusion latency. The useful AVs are those that add evidence in regions where the RSU is uncertain. Conductor exploits this insight by using detector confidence maps that are already produced by the fusion model. It scores each AV by how much its evidence complements the RSU's view, then selects a budget-feasible subset.

Building on that insight, Conductor employs two core ideas. First, instead of fusing inputs from every available AV, it selects the most valuable subset adding the most information over the RSU's view of a locale. Second, it selectively adapts trajectory prediction work on each update, based on remaining time budget. Risky, fast-changing tracks receive fresh learned prediction, while the rest reuse older cached predictions.

Conductor manages the selection and prediction policies and their allocated time budgets jointly, as they collectively dictate the resulting AoI. 
Fusing more AVs can improve perception, but it leaves less time for trajectory prediction. Predicting more tracks can improve planning, but only if the world model still reaches the planner on time. Conductor therefore treats AV selection and trajectory prediction as one scheduling problem under a unified AoI bound.

This paper makes five contributions:
\begin{enumerate}
  \item \textbf{WorldFusion:} an RSU-anchored fusion layer that builds one shared scene for the \textit{locale}. Prior systems either build a fused view per CAV, or use early fusion with raw sensor data. The former repeats computation and requires per-CAV transmission; the latter requires large uploads. WorldFusion instead fuses compact intermediate features once at the edge and sends the same model to all CAVs.

  \item \textbf{Conductor:} an edge service that extends WorldFusion with shared tracking state, trajectory prediction, and one world-model update %
  that can be multicast to all affected CAVs in the \textit{locale}.

  \item \textbf{Deadline-cliff characterization:} a measurement study showing that composing fusion, tracking, and prediction on edge hardware causes AoI to exceed the 300\,ms bound at high AV counts.

  \item \textbf{Occlusion-aware AV selection:} a selector that chooses CAVs based on their added detection value over the RSU's view, using detector confidence maps already produced by the fusion model.

  \item \textbf{Deadline-aware runtime control:} a controller that jointly adapts the number of CAVs fused and the amount of fresh trajectory prediction, keeping AoI within the 300\,ms bound while preserving occlusion-recovery benefits.
\end{enumerate}

Across traffic scenarios with up to 31 CAVs, Conductor is the only evaluated policy that consistently meets the AoI bound. In a highly occluded locale where AV choice is critical, selecting four CAVs recovers most of the objects hidden from the RSU, closing 87\% of the gap between random selection and an Oracle. In closed-loop evaluations, late fusion fails in occluded scenarios, using all available CAVs for feature-level fusion fails in dense scenarios, while Conductor preserves cooperative sensing benefits in all cases.

\section{Related Work}
\label{sec:related}

Conductor is not a new cooperative perception model. It is an edge service for cooperative motion prediction: cooperative perception is its front end, and its output is tracks and predicted trajectories for every planner in a \textit{locale}. \Cref{tab:related} positions the closest systems by input and output. Our work lies at the intersection of cooperative perception, cooperative motion prediction, and latency-aware edge scheduling; prior work improves individual components or composes them without an end-to-end deadline at high actor counts.

\begin{table}[t]
\caption{Task boundaries of the closest systems.}
\label{tab:related}
\footnotesize
\centering
\setlength{\tabcolsep}{3pt}
\begin{tabular}{@{}>{\raggedright\arraybackslash}p{0.37\columnwidth}|>{\raggedright\arraybackslash}p{0.26\columnwidth}|>{\raggedright\arraybackslash}p{0.32\columnwidth}@{}}
\toprule
\textbf{System} & \textbf{Input} & \textbf{Output} \\
\midrule
LiveMap~\cite{livemap2021}, AdaMap~\cite{adamap2023} & \multirow{2}{*}{detected objects} & \multirow{2}{*}{object map, tracks} \\
\hline
\multirow{2}{0.35\columnwidth}{Harbor~\cite{harbor}, C-MASS~\cite{cmass2025}} & selected perception\ data & \multirow{2}{*}{fused detections} \\
\hline
Where2Comm~\cite{hu2022where2comm} & feature regions & detections, one AV \\
\hline
CMP~\cite{cmp}, V2XPnP~\cite{v2xpnp2025}, Co-MTP~\cite{comtp2025} & \multirow{2}{*}{features, tracks} & \multirow{2}{*}{trajectories, one AV} \\
\hline
\multirow{2}{*}{\textbf{Conductor}} & \multirow{2}{0.26\columnwidth}{BEV features (vehicles + RSU)} & tracks + trajectories, all planners in \textit{locale} \\
\bottomrule
\end{tabular}
\end{table}

\subsection{Cooperative Perception}

\textbf{Infrastructure-less vehicle-to-vehicle (V2V)} systems (AutoCast~\cite{Autocast}, Select\-2Col~\cite{select2col2023}, IMAS2~\cite{imas2_2024}, mmCooper~\cite{mmcooper2024}) avoid centralized compute, but duplicate fusion and tracking on each participating CAV. They distribute work across vehicles, while edge-assisted systems centralize work and must ensure that the work is completed within the AoI time constraint. Existing V2V selection methods optimize perception quality at a fixed contributor count, not scalable joint fusion and prediction within an AoI constraint.

\textbf{Edge-assisted vehicle-to-infrastructure (V2I)} systems (EMP~\cite{EMP}, VIPS~\cite{vips2022}, Harbor~\cite{harbor}, F-Cooper~\cite{fcooper}, EdgeCooper~\cite{edgecooper2023}) establish that infrastructure-assisted perception can extend visibility beyond a single vehicle. Edge map systems merge after local detection: LiveMap~\cite{livemap2021} maintains a global object map from detected-object records, and AdaMap~\cite{adamap2023} scales object-level map maintenance to 150-vehicle traces at 70\,ms average end-to-end latency; their latency numbers end before motion prediction and planner consumption, which are the stages Conductor adds and controls. C-MASS~\cite{cmass2025} schedules which sensors contribute to collaborative perception under mobility, complementary to our compute-deadline admission. EdgeCooper~\cite{edgecooper2023} is the closest to our system on communication behavior. It schedules complementary raw LiDAR point clouds over a multi-hop V2X network to construct a holistic edge view. Harbor~\cite{harbor} and related systems optimize bandwidth and delay across V2V and V2I links for cooperative perception. 
None of these systems evaluates the full fusion, tracking, and prediction pipeline at scale nor do they model how their systems adhere to a strict AoI time constraint as our work does.

\textbf{Cooperative perception models} (Where2Comm~\cite{hu2022where2comm}, V2X-ViT~\cite{v2xvit2022}, CoBEVT~\cite{xu2022cobevt}, CoAlign~\cite{coalign2023}) focus on improving detection accuracy through feature-level (intermediate) fusion. These models use per-CAV confidence maps to guide feature aggregation or communication.
Conductor uses these confidence maps for a different purpose: selecting which CAVs add the most value and should participate in fusion given a compute deadline. The selector operates at the edge controller and targets AoI compliance rather than communication efficiency or detection accuracy alone.

\textbf{Fusion approaches.} Cooperative perception systems differ by what they share. Early-fusion systems share raw sensor data, as in raw-LiDAR edge perception systems such as EdgeCooper~\cite{edgecooper2023}. Feature-level (intermediate) systems such as F-Cooper~\cite{fcooper}, Where2Comm~\cite{hu2022where2comm}, V2X-ViT~\cite{v2xvit2022}, and CoBEVT~\cite{xu2022cobevt} share learned representations. Late-fusion systems share perception outputs such as detected objects, boxes, or tracks; AutoCast~\cite{Autocast} is a systems example of object-level V2V sharing, and LFF-V2V~\cite{lff_v2v} is a recent explicit late-fusion framework. Late fusion is bandwidth-efficient, but it cannot recover objects that no individual sensor detects confidently. Conductor uses feature-level fusion, which preserves spatial evidence while avoiding raw point-cloud uploads.

\subsection{Cooperative Motion Prediction}

CMP~\cite{cmp}, the closest prior system, composes CoBEVT cooperative perception, AB3DMOT tracking \cite{ab3dmotTracker2020}, and MTR~\cite{shi2022mtr} into one decentralized stack where each CAV runs its own pipeline on OPV2V/V2V4Real with no hard deadline constraint. CMP is important because it shows that combining cooperative perception, tracking, and learned motion prediction improves the information available to an AV planner. Its stack is representative of the current cooperative motion-prediction pipeline: feature-level cooperative perception, AB3DMOT tracking, and MTR prediction. Conductor adopts this class of pipeline but changes where and how it runs: the stack runs once at the edge for the entire \textit{locale}, under an explicit AoI deadline. Recent joint perception-and-prediction models, V2XPnP~\cite{v2xpnp2025} and Co-MTP~\cite{comtp2025}, fuse spatio-temporal features across agents to produce trajectories for one AV at a time; like CMP they optimize accuracy with no deadline. Both CMP and CoPnP~\cite{copnp2024} limit evaluation scale to 2--7 CAVs. We centralize the stack on the edge so fusion is computed once per update and shared across all recipients, avoiding duplicated per-vehicle fusion.
Centralization raises a scheduling problem that CMP avoids through per-vehicle isolation. We evaluate dense \textit{locales} with up to 31 CAVs, exposing a deadline cliff invisible at CMP's evaluation scale, and present a controller with a per-cycle compute budget that jointly performs fusion and prediction.

\subsection{Edge Compute Scheduling for Autonomy}

General-purpose DNN-serving frameworks treat each request independently and optimize throughput or latency per request. Single-vehicle adaptive compute~\cite{li2020streaming,gog2022pylot} adjusts model complexity within one vehicle's budget, but does not address cooperative workloads where many contributors share one edge deadline. In Conductor, the stages are interdependent: fusion cost depends on the number of participating CAVs, tracking cost depends on detection count, and prediction cost depends on track count. Our controller schedules these stages as one edge service and adheres to the AoI constraint rather than using per-stage throughput as the success criterion.

These gaps motivate treating cooperative perception, tracking, and prediction as one edge-scheduled service with an explicit AoI objective.

\section{System Architecture and Design}
\label{sec:architecture}

Conductor is an edge service that builds and shares a world model for one RSU-anchored \textit{locale}. The world model is the service's output: a timestamped, world-coordinate scene containing persistent 3D tracks, motion state, freshness metadata, and six-mode predicted trajectories, each fresh, cached, or replaced by a constant-velocity fallback. It is not a learned simulator or a generative model. A \textit{locale} contains $N$ cooperating agents in total: one RSU (always included) and up to $N{-}1$ CAVs.
\Cref{tab:notation} summarizes the notations we use in this section for ease of reference.

\begin{table}[t]
\centering
\vspace{4mm}
\caption{Key terms and notation.}
\label{tab:notation}
\footnotesize
\setlength{\tabcolsep}{4pt}
\renewcommand{\arraystretch}{0.95}
\begin{tabular}{ll}
\toprule
Term & Definition \\
\midrule
Locale & RSU-anchored intersection or road segment \\
Contributor & CAV that uploads features to the edge \\
Recipient & CAV that consumes the published world model \\
AoI & World model's age when the CAV planner reads it \\
$N$ & Cooperating agents in a \textit{locale}, including the RSU \\ 
$K_{\mathrm{cav}}$ & Selected CAV contributors for one publish cycle \\
$K$ & Contributors entering fusion, $K=K_{\mathrm{cav}}+1\,RSU$ \\
$E$ & Recipient CAVs consuming the published world model \\
$M$ & Tracked objects in the current update \\
$M_{\mathrm{prev}}$ & Tracked objects from the previous update \\
\bottomrule
\end{tabular}
\end{table}

\begin{figure*}[t]
    \centering
    \usetikzlibrary{positioning, arrows.meta, fit, backgrounds}
    \begin{tikzpicture}[
      font=\footnotesize,
      >=Stealth,
      node distance=3mm and 5mm,
      stage/.style={rectangle, rounded corners=2pt, draw=black!70,
                    fill=white, minimum height=8mm, minimum width=18mm,
                    align=center, inner sep=1.5pt, font=\scriptsize},
      ctrlnode/.style={rectangle, rounded corners=2pt, draw=black!70,
                       dashed, fill=white, minimum height=8mm,
                       minimum width=28mm, align=center, inner sep=2pt,
                       font=\scriptsize},
      grouplabel/.style={font=\scriptsize\itshape, text=black!75},
      netlabel/.style={font=\scriptsize\itshape, text=black!75},
      arr/.style={->, semithick, draw=black!80},
      ctrlarr/.style={->, dashed, draw=black!60},
    ]

    \node[stage] (sense) {Sensing\\(LiDAR)};
    \node[stage, right=of sense] (vox) {Voxelize};
    \node[stage, right=of vox]   (bb)  {Backbone\\(PointPillars)};
    \node[stage, right=of bb]    (psm) {Per-agent\\PSM head};
    \node[stage, right=of psm]   (comp) {Compress\\$\sim$17\,KB};
    \draw[arr] (sense) -- (vox);
    \draw[arr] (vox)   -- (bb);
    \draw[arr] (bb)    -- (psm);
    \draw[arr] (psm)   -- (comp);

    \node[stage, below=12mm of sense] (filt) {Occlusion-aware\\filter (PSM)};
    \node[stage, right=of filt] (fuse) {Cross-attn\\fusion};
    \node[stage, right=of fuse] (det)  {Detection};
    \node[stage, right=of det]  (trk)  {Tracking\\(AB3DMOT)};
    \node[stage, right=of trk]  (pred) {Prediction\\(MTR)};
    \node[stage, right=of pred] (sserw) {Serialize\\$\sim$13\,KB};

    \draw[arr] (filt) -- (fuse);
    \draw[arr] (fuse) -- (det);
    \draw[arr] (det)  -- (trk);
    \draw[arr] (trk)  -- (pred);
    \draw[arr] (pred) -- (sserw);

    \node[stage, above=12mm of sserw] (deserw) {Deserialize};
    \node[stage, right=of deserw] (plan) {Planner\\(20\,Hz)};
    \draw[arr] (deserw) -- (plan);

    \coordinate (ul_mid) at ($(comp.south) + (0, -8mm)$);
    \draw[arr, rounded corners=3pt]
      (comp.south) -- (ul_mid) -| (filt.north);
    \node[netlabel, right=1mm of ul_mid, anchor=west, align=left]
      {V2X uplink (Uu)\\$N$ parallel streams};

    \coordinate (dl_mid) at ($(sserw.north) + (0, 8mm)$);
    \draw[arr, rounded corners=3pt]
      (sserw.north) -- (dl_mid) -| (deserw.south);
    \node[netlabel, right=1mm of dl_mid, anchor=west, align=left]
      {V2X downlink\\(NR-MBS multicast)};

    \begin{scope}[on background layer]
      \node[draw=black!40, rounded corners=3pt, fill=blue!3,
            fit=(sense) (vox) (bb) (psm) (comp),
            inner sep=4pt,
            label={[grouplabel, anchor=west]above left:%
              Contributor CAVs, in parallel}] {};
      \node[draw=black!40, rounded corners=3pt, fill=black!3,
            fit=(filt) (fuse) (det) (trk) (pred) (sserw),
            inner sep=4pt,
            label={[grouplabel, anchor=west]above left:%
              Edge (RSU-anchored, one per \emph{locale})}] {};
      \node[draw=black!40, rounded corners=3pt, fill=green!3,
            fit=(deserw) (plan),
            inner sep=4pt,
            label={[grouplabel, anchor=west]above left:%
              Recipient vehicle}] {};
    \end{scope}

    \node[ctrlnode, below=8mm of fuse] (csel)
      {Contributor selection\\($K_{\mathrm{cav}}$ adaptive)};
    \node[ctrlnode, below=8mm of trk] (amort)
      {Prediction\\amortization\\(divergence cache)};
    \node[ctrlnode, below=8mm of pred] (sched)
      {Risk-budgeted\\MTR scheduling};

    \draw[ctrlarr] (csel.north)  -- (filt.south);
    \draw[ctrlarr] (amort.north) -- (trk.south);
    \draw[ctrlarr] (sched.north) -- (pred.south);

    \begin{scope}[on background layer]
      \node[draw=black!50, dashed, rounded corners=3pt, fill=white,
            fit=(csel) (amort) (sched),
            inner sep=4pt,
            label={[grouplabel, anchor=west]below left:%
              Controller (compute budget $D$)}] {};
    \end{scope}

    \end{tikzpicture}
    \vspace{-2mm}
    \caption{Conductor pipeline. Contributors uplink compressed features, confidence maps, and poses; the edge filters to $K_{\mathrm{cav}}$, runs fusion, detection, tracking, and prediction once per update, and multicasts the world model. Recipients consume it on their 20\,Hz planner cycle. The controller (bottom) adapts $K_{\mathrm{cav}}$ and prediction scheduling under the compute SLO $D$. %
    }
    \label{fig:system_overview}
\end{figure*}
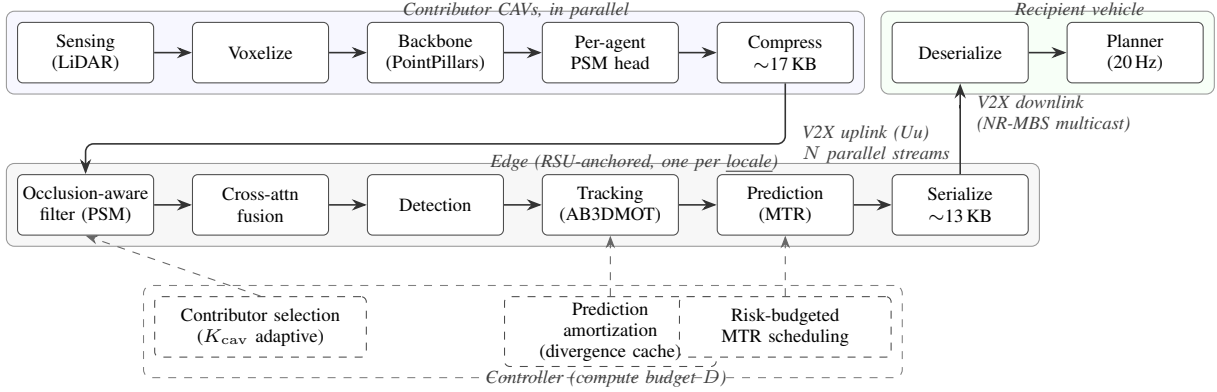

\Cref{fig:system_overview} shows Conductor's end-to-end pipeline. On each update, the edge collects compressed features from the \textit{locale}'s RSU and CAVs, selects the RSU and $K_{\mathrm{cav}}\leq N-1$ CAVs, fuses their features, updates object tracks, and predicts short-horizon trajectories.
Conductor then sends one shared world-model update to every CAV in the \textit{locale}, using one multicast transmission per update. 

Each pipeline stage adds information the AV planner ultimately uses. Feature-level fusion to create a unified world view recovers objects that no single CAV or RSU can detect confidently. Tracking gives those detections persistent identity across updates. Motion prediction turns tracked objects into future trajectories. Conductor is therefore not only a perception service, but a tracking and trajectory prediction service whose output is useful to planners if delivered within the AoI constraint.

This section describes Conductor's design requirements (\cref{ssec:requirements}), service boundary (\cref{ssec:service-boundary}), RSU-anchored fusion (\cref{ssec:rsu-fusion}), edge pipeline (\cref{ssec:edge-pipeline}), and runtime knobs (\cref{ssec:control-knobs}). \Cref{sec:controller} then describes the controller that operates those knobs.

\subsection{Design Principles}
\label{ssec:requirements}

Three principles shape Conductor's architecture:

\begin{enumerate}
  \item \textbf{One shared world model per \textit{locale}.} Construction of a separate fused view for each receiving CAV repeats similar edge computation and requires a separate downlink transmission per CAV. Instead, Conductor fuses selected CAV features \textit{once} with the RSU features and multicasts the same world-model update to every CAV in the \textit{locale}.
  \item \textbf{Persistent state across updates, anchored at the RSU.} A shared coordinate anchor keeps track identities consistent across updates as CAVs move through the \textit{locale}. The RSU embodies that spatial anchor, so object tracks and trajectory predictions retain stable identity from one update to the next.
  \item \textbf{Deliver each update within the AoI bound.} The world model is useful only if it reaches CAV planners while still fresh. We use 300\,ms as the upper bound on AoI---derived in \cref{sec:intro} from the braking margin at urban speed (4\,m of travel, one lane width, at 14\,m/s) and validated by the closed-loop conflict experiments in \cref{ssec:closed_loop}---and design edge pipeline intelligence so its compute share remains bounded as $N$ grows.
\end{enumerate}

\subsection{Service Boundary and Multicast Scaling}
\label{ssec:service-boundary}

Each CAV continuously uploads its compressed feature output to the edge over its own C-V2X 5G Uu uplink. The edge's input is therefore a set of (feature tensor, per-CAV confidence map, pose) tuples---one per CAV plus one from the RSU. The pose carries each RSU/CAV's position and heading in the shared frame. The edge's output is one world-model update per cycle, containing the current tracks and short-horizon trajectory predictions for all objects in the \textit{locale}.

Because the output is identical for every receiving CAV, the edge can multicast the update once on the cellular downlink rather than scheduling $E$ separate transmissions, one per recipient CAV. This multicast property follows from the shared world model: every CAV consumes the same RSU-anchored state. \Cref{ssec:network} quantifies the radio model.

\subsection{RSU-Anchored Intermediate Fusion}
\label{ssec:rsu-fusion}

Conductor uses intermediate fusion because early fusion is bandwidth-bound and late fusion discards spatial structure. Early fusion ships raw point clouds, on the order of 10\,MB per RSU/CAV per update, exceeding V2X uplink capacity at moderate $N$. Late fusion compresses payloads but delivers only object lists, leaving the edge unable to reason about occluded regions or to combine partial detections from multiple CAVs into one confident detection. Intermediate fusion preserves the spatial structure and compresses post-encoder features to roughly 17\,KB per CAV, fitting practical V2X uplink capacities.

Anchoring fusion in the RSU frame serves two purposes. First, it gives the \textit{locale} one persistent coordinate frame, so tracks and predictions remain consistent across updates as CAVs move through the intersection. Second, the RSU is a structurally privileged sensor: its traffic-light-mounted LiDAR at 4--6\,m elevation covers most of the intersection center, reducing the need for contributions from CAVs that duplicate its view.

In our empirical findings, the RSU alone recovers 54\% of in-\textit{locale} detections, and adding all $K_{\mathrm{cav}}{=}22$ CAVs raises recall to only 60\% at $4{\times}$ the fusion cost. The 60\% ceiling reflects the detector's intrinsic recall on Multi-V2X (AP@0.5${=}0.674$, see \cref{sec:discussion}); the remaining objects lie outside every contributor's range or are jointly occluded and cannot be recovered by any fusion. Not all CAVs improve recall equally, as many duplicate the RSU's view. Conductor therefore always includes the RSU and selects only the CAVs that add evidence outside the RSU's view. \Cref{sec:controller} details the selection logic.

\subsection{Edge Pipeline}
\label{ssec:edge-pipeline}

\Cref{fig:system_overview} depicts the three core elements of the pipeline: contributing CAV feature extraction (top left), edge fusion and prediction (middle), and receiving CAV planner consumption (top right), with one network hop in each direction. The bottom block shows the controller (\cref{sec:controller}).

\textbf{Contributing CAV} (\cref{fig:system_overview}, top left). Each CAV voxelizes its raw LiDAR into a local pillar grid, runs a PointPillars backbone~\cite{lang2019pointpillars} to produce a compressed feature tensor (a bird's-eye-view representation), runs a per-RSU/CAV classification head that produces a per-RSU/CAV spatial confidence map (PSM), and transmits the (feature tensor, PSM, pose) tuple over its uplink.
Every CAV does this in parallel, so the latency the edge sees is set by the slowest CAV.

\textbf{Edge} (\cref{fig:system_overview}, middle). Incoming packets are inserted into a bounded jitter buffer keyed by source timestamp, so the downstream tracker sees temporally ordered inputs even when the uplink reorders or delays packets. At each update, the edge carries out the following steps:

\begin{enumerate}
  \item \textbf{Selects} $K_{\mathrm{cav}}$ CAVs with the occlusion-aware filter (\cref{ssec:selector}), using the per-RSU/CAV PSMs already in the input. The filter requires no additional model inference.
  \item \textbf{Runs} WorldFusion's Where2Comm-style cross-attention fusion~\cite{hu2022where2comm}, warping the selected feature tensors into the RSU frame and reconciling them under the per-RSU/CAV PSMs.
  \item \textbf{Detects} objects on the fused tensor and emits world-frame 3D boxes.
  \item \textbf{Updates} a vectorized AB3DMOT tracker~\cite{ab3dmotTracker2020} to maintain \textit{locale}-level identity across updates. The stock tracker runs predict, gating, and update sequentially per track, so its per-update cost grows linearly in track count. We replaced this with a batched Kalman update over all active tracks, giving a 4--5$\times$ speedup at our evaluated deployment scale.
  \item \textbf{Runs} the MTR predictor~\cite{shi2022mtr} over the fused tensor to produce multi-modal trajectory predictions.
  \item \textbf{Multicasts} the world-model update ($\sim$13\,KB) on the cellular downlink.
\end{enumerate}

\textbf{Receiving CAV} (\cref{fig:system_overview}, top right). Each receiving CAV deserializes the most recent world-model update and the planner reads it on its 50\,ms control cycle. Every CAV receives the same bytes because the world model is anchored in the shared RSU frame, so the per-recipient downlink scheduler does not grow with the number of receiving CAVs.

\subsection{Runtime Control Knobs}
\label{ssec:control-knobs}

The cost of each pipeline component scales differently. CAV-side cost is bounded by single-CAV compute and is insensitive to $N$ because CAVs run their perception in parallel. Edge-side cost grows with $K_{\mathrm{cav}}$. Cross-attention fusion dominates at moderate-to-high CAV counts and drives the deadline cliff that motivates our work.

To stay within the AoI bound, the edge targets a 130\,ms compute Service-Level Objective (SLO), which is the residual budget after subtracting calibrated network delay and the planner wait time (\cref{ssec:network}) from the 300\,ms AoI bound at high $N$. Two quantities are dynamically adapted on each update:
\begin{itemize}
  \item $K_{\mathrm{cav}}$, the number of CAVs selected to participate in fusion.
  \item $\mathbf{x}$, the set of tracked objects that receive fresh learned prediction; the rest reuse a cached prediction or fall back to a lightweight linear extrapolator.
\end{itemize}
The two share one compute budget: fusing more CAVs improves scene quality but leaves less budget for prediction, while more track predictions help the planner only if information reception is timely. The controller (\cref{sec:controller}) formulates this set of constraints as a joint optimization over $K_{\mathrm{cav}}$ and $\mathbf{x}$ under a per-stage cost model fit on profiled hardware, selecting the pair that maximizes estimated prediction utility while keeping the edge within the 130\,ms compute SLO. \Cref{sec:eval} reports the measured deadline cliff and the controller's response across CAV density, dictated by the participant count $N$.

\section{Adaptive Controller}
\label{sec:controller}

\Cref{sec:architecture} defines Conductor's dataflow and two runtime knobs: how many CAVs are fused and which tracks receive fresh prediction. This section gives the controller that operates those knobs. The controller extends the system architecture by assigning the per-update edge budget between CAV selection, fusion, and trajectory creation so that each world-model update remains within the AoI bound.

On each update, the controller decides (i) how many CAV contributors are chosen for fusion, and (ii) which tracked objects receive fresh learned prediction (\cref{alg:controller}). We denote these by $K_{\mathrm{cav}}$ and $\mathbf{x} \in \{0,1\}^M$, where $x_i{=}1$ means object $i$ receives fresh MTR inference and $x_i{=}0$ means it reuses a cached prediction or falls back to the lightweight linear predictor. The per-cycle edge cost is:
\begin{equation}
\begin{split}
  C(K, \mathbf{x}) = {} & c_{\mathrm{filter}}(N, M_{\mathrm{prev}}) + c_{\mathrm{fuse}}(K) + c_{\mathrm{det}} \\
    & + c_{\mathrm{track}}(M)
    + c_{\mathrm{pred}}(\mathbf{x}) + c_{\mathrm{oh}}
\end{split}
  \label{eq:cost}
\end{equation}
where $C(K,\mathbf{x})$ is the estimated per-update edge compute cost; $K = 1 + K_{\mathrm{cav}}$ is the RSU plus the $K_{\mathrm{cav}}$ selected CAVs whose features are fused; $N$ is the cooperating RSU + CAVs at the current update; $M$ is the estimated tracked-object count for the current update; and $M_{\mathrm{prev}}$ is the tracked-object count from the previous update. $c_{\mathrm{filter}}$, $c_{\mathrm{fuse}}$, $c_{\mathrm{det}}$, $c_{\mathrm{track}}$, and $c_{\mathrm{pred}}$ are the per-stage compute costs of CAV selection, fusion, detection, tracking, and learned prediction, respectively. $c_{\mathrm{oh}}$ is a small fixed per-cycle overhead covering controller bookkeeping and pipeline glue outside the main model stages.

The controller needs a fast latency estimate before choosing $K_{\mathrm{cav}}$ and $\mathbf{x}$. We profile the pipeline once on the deployment GPU and fit a simple model for each stage: fusion depends on the number of fused contributors, tracking depends on the number of tracked objects, and prediction depends on the number of objects selected for fresh prediction inference. At runtime, $N$ and $M_{\mathrm{prev}}$ are known before scheduling begins. The controller estimates the current track count $M$ from the selected contributor count using a 20-update rolling tracks-per-contributor ratio $\bar{\rho}$ plus a slack term, $\hat{M} = \bar{\rho}K + \mathrm{slack}$ with $\mathrm{slack} = \max(0.1\,\bar{\rho}K,\, 2)$ tracks.

Three mechanisms control $K_{\mathrm{cav}}$ and $\mathbf{x}$:

\subsection{Mechanism 1: Occlusion-Aware Contributor Selection}
\label{ssec:selector}

The RSU is always included due to its superior vantage point.
When the $N{-}1$ candidate CAVs exceed the capacity-feasible count, the controller selects $K_{\mathrm{cav}} < N{-}1$ CAVs whose detector evidence most complements the RSU's view.

Each contributor (RSU and each CAV) also produces a spatial confidence map (PSM), following Where2Comm~\cite{hu2022where2comm}. A high value in a grid cell means the contributor is confident that an object is present. The PSM is already present for fusion (uploaded alongside the feature tensor), so the selector can use it without adding another inference.

Let $\psi_j(u,v) \in [0,1]$ denote CAV $j$'s PSM evaluated at cell $(u,v)$ of the feature-tensor grid, and $\psi_{\mathrm{rsu}}(u,v)$ the RSU's PSM at the same cell. The controller scores each candidate CAV $j$ by
\begin{equation}
  s_j = \sum_{(u,v)} \psi_j(u,v) \cdot \bigl(1 - \psi_{\mathrm{rsu}}(u,v)\bigr)
  \label{eq:contrib}
\end{equation}
where $s_j$ is CAV $j$'s selection score, summed over all cells of the feature-tensor grid. The score is high when CAV~$j$ detects an object that the RSU does not. A distance gate drops CAVs beyond a configurable range (50\,m by default).

The score is a dot product over the feature-tensor grid. Edge-side scoring cost is dominated by the dot-product reduction. The per-RSU/CAV classifier outputs are already part of the fusion path's input, so the controller adds no extra inference.

The controller searches candidate values $K_{\mathrm{cav}} \in \{0, 2, 4, 8, 12\}$ and selects the $K^*_{\mathrm{cav}}$ that maximizes estimated utility given the cost model.
Estimated utility at a given $K_{\mathrm{cav}}$ is the number of tracks expected to receive fresh predictions within the remaining prediction budget. \cref{sec:mech2} defines how the controller decides which cached predictions need refresh.

\begin{algorithm}[tb]
\small
\caption{Controller actions per update.}
\label{alg:controller}
\begin{algorithmic}[1]
\Require RSU + CAV features for $N$ cooperating units, deadline $D$
\State Score CAVs by added evidence outside the RSU view (\cref{eq:contrib})
\ForAll{$K_{\mathrm{cav}} \in \{0, 2, 4, 8, 12\}$}
  \State $\hat{M} \gets \bar{\rho} K + \mathrm{slack}$
  \State $\hat{C} \gets c_{\mathrm{filt}} + c_{\mathrm{fuse}}(K)
    + c_{\mathrm{det}} + c_{\mathrm{track}}(\hat{M})$
  \If{$\hat{C} + c_{\mathrm{oh}} > D$} \textbf{skip} \EndIf
  \State $B \gets D - \hat{C} - c_{\mathrm{oh}}$
  \State Estimate utility from $B$ and divergence rate
\EndFor
\State $K^* \gets \arg\max_{K} U(K, B)$
\State Fuse the RSU and top-$K^*_{\mathrm{cav}}$ CAVs; detect; track
\State Gate divergence (\cref{eq:divergence})
\State $B_{\mathrm{res}} \gets D - C_{\mathrm{measured}}$
\State Rank divergent objects by $r_i$ (\cref{eq:risk})
\State Assign $x_i{=}1$ greedily within $B_{\mathrm{res}}$
\State Run MTR on selected tracks; use the lightweight linear predictor for the rest
\State \Return updated world model
\end{algorithmic}
\end{algorithm}

\subsection{Mechanism 2: Prediction Cost Amortization}
\label{sec:mech2}
\label{ssec:amortization}

Not every tracked object requires fresh prediction each cycle.
The controller maintains a prediction cache keyed by track~ID.
Object~$i$ is flagged \emph{divergent} if:
\begin{equation}
\begin{split}
  \|\hat{p}_i(t) {-} p_i^{\mathrm{obs}}(t)\| &>
    \max(d_{\mathrm{floor}},\, v_i \tau) \\
  \text{or} \quad
  |\mathrm{wrap}(\hat{\theta}_i(t) {-} \theta_i^{\mathrm{obs}}(t))| &>
    \theta_{\mathrm{thr}}
\end{split}
  \label{eq:divergence}
\end{equation}
where $\hat{p}_i(t)$ is object $i$'s cached position prediction evaluated at the current update, $p_i^{\mathrm{obs}}(t)$ is the Kalman-filtered observed position, $\|\cdot\|$ is Euclidean distance, $\hat{\theta}_i(t)$ and $\theta_i^{\mathrm{obs}}(t)$ are the corresponding cached and observed heading angles, $\mathrm{wrap}(\cdot)$ maps angular difference to $[-\pi, \pi]$ so divergence does not fire spuriously across the $\pm180^{\circ}$ boundary, $v_i$ is object speed, $\tau = 0.1\,$s is the time-domain tolerance, $d_{\mathrm{floor}} = 0.5\,$m is a minimum-displacement floor that prevents triggering on stationary tracking noise, and $\theta_{\mathrm{thr}} \approx 10^{\circ}$ is the heading-divergence threshold.
The velocity-normalized threshold allows up to 100\,ms of
unmodeled drift regardless of speed. A max cache age forces refresh even for non-divergent objects.

\subsection{Mechanism 3: Risk-Budgeted Scheduling}
\label{ssec:risk}

Among divergent objects, the controller allocates the residual prediction budget to those with highest risk:
\begin{equation}
\begin{split}
  r_i = v_i \cdot \exp\!\bigl({-}d_i / d_{\mathrm{scale}}\bigr)
    \cdot \omega_i \\
    \cdot \max\!\bigl(0,\; 1 {-} c_i / r_{\mathrm{safe}}\bigr)
\end{split}
  \label{eq:risk}
\end{equation}
where $r_i$ is object $i$'s risk score, $v_i$ is its speed, $d_i$ is its distance to the nearest CAV, $d_{\mathrm{scale}} = 20\,$m sets the spatial decay scale, $\omega_i \in \{1, 2\}$ is an occlusion weight (2 if the object lies in an RSU blind cell), $c_i$ is the minimum constant-velocity closest-approach distance between object $i$ and every other tracked CAV over the 5\,s horizon (excluding self-comparison when $i$ is itself a CAV; no planner path is uploaded), and $r_{\mathrm{safe}} = 5\,$m is the safe-clearance distance below which the closest-approach factor saturates. The score is a scheduling heuristic for prioritizing fresh prediction inference, not a verified collision-risk estimate.

The scheduler ranks divergent objects by $r_i$ and greedily assigns $x_i {=} 1$ until the residual prediction budget
$B_{\mathrm{pred}} = D - C_{\mathrm{measured}}$ is exhausted.
Objects that do not receive fresh MTR inference fall back to a lightweight linear predictor, which extrapolates their future motion from their recent tracked state.

\section{Methodology}
\label{sec:method}

\subsection{Experimental Platform}
\textbf{Two evaluation platforms.} We conduct our evaluations on two complementary platforms. Both share the same edge stack, the same A10 hardware, and the same ns-3 5G NR co-simulation for the network model. They differ in the data source feeding the pipeline.

The \emph{offline profiler} platform replays Multi-V2X~\cite{multiv2x2024} point clouds at every \textit{locale}, runs the WorldFusion + AB3DMOT + MTR pipeline end-to-end on the A10, and composes ns-3 co-simulation uplink samples and analytical NR-MBS downlink with the measured edge compute to produce AoI distributions. The offline profiler platform is used in \cref{ssec:cliff,ssec:cross_scale,ssec:contributor_selection,ssec:joint_controller,ssec:trk_pred_quality,ssec:aoi}.

The \emph{closed-loop} platform runs the full eCAV stack~\cite{eCAV}, an edge-assisted CAV simulation platform that orchestrates CARLA, distributed CAV actors, an edge service container, and ns-3 5G-LENA co-simulation. %
CARLA 0.9.15 (synchronous, $\Delta t = 50\,$ms) simulates the scenario, the edge service runs in its production container as in the offline profiler platform, and ns-3 mediates uplink and downlink live, in step with the simulator clock. This platform is used in \cref{ssec:closed_loop}.

\textbf{Hardware.} Edge components run on an NVIDIA A10 (24\,GB VRAM) deployed as an Azure NV36ads~A10~v5 edge node. The A10 is representative of edge-class accelerators deployed at cell-tower edge sites. A single edge server may serve multiple \textit{locales}; the effective per-\textit{locale} budget is a fraction of the server's total capacity.

\textbf{Cadence.} The planner control loop runs at 20\,Hz. Each CAV planner consumes the latest published world model every $\Delta t {=} 50$\,ms. The cooperative perception path runs at a lower cadence: WorldFusion fires every 100\,ms (10\,Hz) under the Conductor adaptive edge configuration. The late-fusion baseline publishes once per 50\,ms planner cycle, since per-agent 3D detection and the object-level merge are both cheap.

\subsection{Dataset}
We evaluate on Multi-V2X~\cite{multiv2x2024}, a CARLA-based cooperative perception dataset with 6 towns, 56 RSU-anchored \textit{locales}, and dynamic agent connectivity.
Each \textit{locale} is a signalized intersection with one RSU mounted at 4--6\,m elevation on the traffic light, with its own 64-channel LiDAR (1.3M points/s, 120\,m range, FOV $0^{\circ}$ to $-40^{\circ}$). Connected vehicles carry 64-channel LiDAR with the same point density.

We measured the connectivity distribution across all 56 \textit{locales} and 17{,}525 frames. The mean connected agent count is $N_{\mathrm{total}}{=}11.3$ (median: 10, p95: 21, max: 33).
$N{\geq}25$ occurs in 2.6\% of frames. We report results on two \textit{locales} representing two deployment cases. Locale~A (Town05, $\bar{N}{=}23$, RSU-occlusion 9\%) is dense and has low RSU occlusion: the deadline cliff is most severe here, and the RSU already covers most of the scene, so which CAVs are selected has little marginal effect. Locale~B (Town03, $\bar{N}{=}12$, occlusion 45\%) has moderate traffic but wide RSU blind spots, so which CAVs are selected decides whether occluded objects are recovered. Cross-density behavior is reported as an aggregate over the full Multi-V2X dense \textit{locale} set (\cref{ssec:cross_scale}).

\subsection{Participant Scaling}
We evaluate one \textit{locale} at a time. To isolate the effect of contributor count, we sweep
$K_{\mathrm{cav}} \in \{0, 2, 4, 8, N{-}1\}$, where
$K_{\mathrm{cav}}{=}0$ fuses only the RSU and
$K_{\mathrm{cav}}{=}N{-}1$ fuses all CAVs.
The RSU is always included. For each $K_{\mathrm{cav}}$ value, we run 50 frames of steady-state profiling (frames $\geq 5$ after tracker warmup).

\subsection{Baselines and Selectors}
\label{method:config}

We compare seven configurations that isolate each mechanism and their composition. Configurations (i)--(iv) hold the prediction stage fixed and vary contributor handling. Configurations (v)--(vii) layer adaptive prediction onto the contributor-side variants.

\textbf{(i)~Static all.}  Full pipeline with all $N$ agents
fused, no contributor selection, no prediction adaptation. Static all is the baseline that exposes the deadline cliff.

\textbf{(ii)~Random.}  $K_{\mathrm{cav}}$ CAVs selected
uniformly at random, without any scoring. Any intelligent selector must outperform random to justify its compute cost.

\textbf{(iii)~Oracle.}  Offline greedy forward selection that runs full fusion/detection for each candidate and selects the CAV with the largest marginal correct-detection gain. It is not deployable and serves only as an upper bound.

\textbf{(iv)~Occlusion-aware.}  \textit{Fixed}-$K_{\mathrm{cav}}$ selection using \cref{eq:contrib}. This configuration isolates selector quality without the budget-adaptive choice of $K_{\mathrm{cav}}$.

\textbf{(v)~Filter-only.}  Mechanism 1 only: the Occlusion-aware selector, but with \textit{budget-adaptive} $K^*_{\mathrm{cav}}$ selection, and static prediction (no Mechanisms 2+3).

\textbf{(vi)~Prediction-only.}  Mechanisms 2+3 only: static fusion of all contributors paired with the adaptive predictor.

\textbf{(vii)~Joint controller.}  All three mechanisms.

\smallskip

\noindent Configuration (vii) represents Conductor's fully equipped controller, while 
(iv)--(vi) all leverage subsets of the Joint controller's mechanisms to isolate the contribution of each.

\subsection{Metrics}
We report three classes of metrics.

\textbf{Systems metrics.}  Per-stage latency (filter, fusion, detection, tracking, prediction), total edge latency (mean and p95), and AoI envelope compliance (fraction of updates whose AoI stays inside the 300\,ms safety bound).
We also report how often edge compute alone exceeds the controller's 130\,ms internal service-level objective (SLO). The SLO is derived from the 300\,ms safety bound by reserving time for network delay and planner consume lag, so it is a controller target rather than the paper's top-line safety claim.

\textbf{Detection quality.}  Total recall and precision against
ground truth. We use a 1.5\,m center distance matching threshold (under one car-width); a wider threshold credits a detection of a visible vehicle as a true positive for an occluded vehicle parked behind it, which inflates RSU-only occluded recall artifactually. We also report \emph{RSU-occluded recall}: recall restricted to objects whose 3D line-of-sight from the RSU is blocked by another tracked object. RSU-occluded recall isolates the value of cooperative contributors beyond the RSU baseline.

\textbf{Tracking quality.}  MOTA (Multi-Object Tracking Accuracy)
and ID switch count per sequence.

\subsection{AoI Model and Thresholds}
\label{ssec:slo_derivation}
\label{ssec:network}

\textbf{Thresholds.}  Two thresholds appear repeatedly in the paper. The first is a 300\,ms safety bound on AoI, the largest world-model age for which an RSS-style~\cite{rss} emergency brake still completes before the focal CAV enters the conflict zone. At 14\,m/s (50\,km/h) urban approach speed, a 300\,ms age consumes about 4\,m of reserve toward a crossing actor, roughly one lane width. We confirm this bound in closed loop in \cref{sssec:sop}: sweeping a single delivery delay on an Oracle baseline, success holds $\approx 1.0$ for $\Delta {\leq} 280$\,ms and drops sharply past 320\,ms.

The second threshold is a 130\,ms internal compute SLO, the controller's edge-side target derived from the 300\,ms bound by accounting for the network and consume terms. At $N{=}24$, uplink, downlink, serialize/deserialize, and consume lag together contribute about 130\,ms at p95, leaving 170\,ms of the 300\,ms bound for edge compute. We set the controller target to 130\,ms (40\,ms below the 170\,ms remainder) so that compute-side jitter does not breach the bound at dense $N$. The 130\,ms SLO is a controller knob, not a safety claim.

\textbf{Network terms.}  Contributor features and the published world model are exchanged over a cellular V2X link. Uplink (UL) uses one Uu bearer per contributor at a ${\sim}17$\,KB compressed feature payload. We calibrate per-($N$, payload) delivery delay against an ns-3 5G NR co-simulation: mean delay grows from 17\,ms at $N{=}4$ to 36\,ms at $N{=}31$ at a packet reception ratio (PRR) ${\geq}0.97$, where PRR is the fraction of transmitted packets received within the per-update budget. The controller's filter further shrinks the active uplink count from $N{-}1$ to $K_{\mathrm{cav}}$.

Downlink (DL) carries the ${\sim}13$\,KB world model. We model it analytically as one 5G NR Multicast/Broadcast Service (NR-MBS) transmission per update (Rel-17~\cite{3gpp_ts23247_mbs}) at ${\sim}9$\,ms regardless of recipient count. Multicast is applicable because the world model is anchored in a single shared frame, so all recipients receive identical bytes. We use an analytical model rather than simulation for two reasons. First, the public ns-3 NR module does not implement NR-MBS. Second, per-recipient Uu unicast saturates the base-station scheduler at dense $E$ and is not a viable alternative.

Three additional AoI terms enter the AoI composition: protobuf serialize and deserialize (${\sim}3$\,ms each direction, calibrated on the A10), edge compute (per update from the offline profiler), and planner consume lag. \Cref{ssec:aoi} composes all six terms per cycle into the end-to-end AoI distribution.

\section{Evaluation}
\label{sec:eval}

This section evaluates Conductor in three steps. First, we characterize the deadline cliff and test whether the Joint controller keeps AoI within the 300\,ms bound across participant density (\cref{ssec:cliff,ssec:cross_scale,ssec:aoi}). Second, we isolate the controller mechanisms: the occlusion-aware selector, the prediction scheduler, and their combined effect on detection, tracking, and prediction quality (\cref{ssec:contributor_selection,ssec:joint_controller,ssec:trk_pred_quality}). Third, we validate that the measured AoI and perception effects translate into closed-loop driving outcomes under occlusion and edge load (\cref{ssec:closed_loop}).

\subsection{Deadline Cliff Characterization}
\label{ssec:cliff}

The deadline cliff is caused by fusion cost: as more CAVs participate, fusion becomes the dominant edge stage and pushes p95 AoI past the 300\,ms bound.

\Cref{fig:cliff_per_stage} (introduced in \cref{sec:intro}) reports per-stage latency for the static pipeline aggregated across the dense \textit{locale} set, with the per-agent backbone attributed to the contributor side. Edge fusion grows linearly in the number of contributors entering fusion: from 10\,ms at $K_{\mathrm{cav}}{=}0$ (RSU only) to 114\,ms at $K_{\mathrm{cav}}{=}22$ on Locale~A.

The latency contribution of other edge stages---edge detection ($\sim$25\,ms), tracking ($\sim$17\,ms), prediction (50--80\,ms)---is largely insensitive to $K_{\mathrm{cav}}$. 
Prediction cost scales with the number of tracked objects in the \textit{locale}, not with $K_{\mathrm{cav}}$, which include both cooperating CAVs and non-cooperating vehicles that the RSU detects. Vehicle backbone runs in parallel across contributors and is bounded by the slowest contributor at $\sim$8\,ms.

Composing edge compute with CAV backbone, ns-3-sampled uplink, multicast downlink, and planner consume lag (uniform on $[0, 50)$\,ms, the wait to the next 50\,ms planner boundary), AoI at p95 crosses the 300\,ms safety bound between $K_{\mathrm{cav}}{=}17$ and $K_{\mathrm{cav}}{=}19$.
The cliff is structural. Where2Comm's pairwise spatial attention is linear in $K$ in practice, so fusion cost grows roughly as $11 + 8K$\,ms. Detection and tracking scale with scene density, not with $K$, so reducing contributor count does not help them. Prediction is bounded by MTR's batched inference over tracked objects and grows only with track count. Fusion is therefore the dominant edge compute term in our setting, and meeting the SLO requires controlling contributor count first.

\subsection{Cross-Scale Controller Behavior}
\label{ssec:cross_scale}

We next test whether controlling fusion and prediction together is enough to avoid the cliff across the full Multi-V2X density distribution.

We run the full edge pipeline across the native Multi-V2X \textit{locale} density distribution ($N \in [4, 32]$) under three configurations (Static, Prediction-only, Joint) and report AoI envelope compliance per $N$ bin. Filter-only is omitted because it matches Joint on AoI compliance in this experiment: both choose the same budget-feasible contributor count, so both reduce fusion to the same feasible cost. Joint differs by also adapting prediction work, which affects prediction quality and residual compute margin rather than the compliance curve in this setting.

\begin{figure}[t]
  \centering
  \includegraphics[width=\columnwidth]{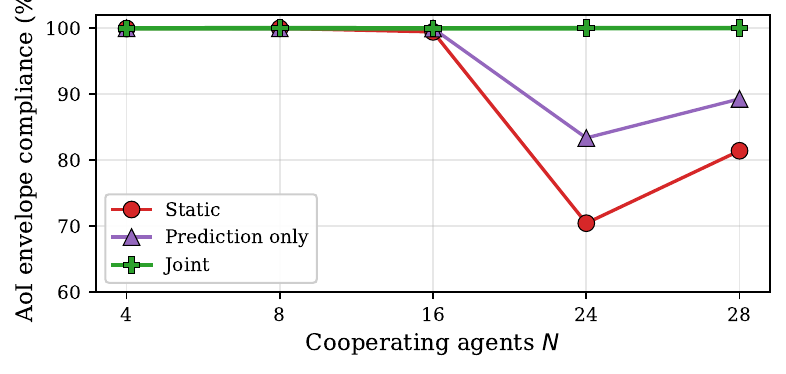}
  \vspace{-7mm}
  \caption{Envelope compliance (\% of updates with composed AoI ${\leq}300$\,ms) vs.\ participant count $N$ for three policies. The Joint controller is the only policy that complies with the 300\,ms safety bound across the full density sweep. Static and Prediction-only both collapse beyond $N{=}16$ as fusion overshoots the deadline.
  }
  \label{fig:cross_scale_compliance}
\end{figure}

\Cref{fig:cross_scale_compliance} shows that the Joint controller is the only policy that holds 100\% AoI compliance across the full density sweep. Static and Prediction-only collapse to 87\% and 84\% at $N{=}24$, where the cost of fusion saturates the deadline. \Cref{fig:cross_scale_compliance} reports compliance only; selector detection quality is discussed next.

\subsection{Contributor Selection Impact}
\label{ssec:contributor_selection}

The selector is useful for two reasons: it reduces compute everywhere, and it improves recall when the RSU has blind spots.

The selector's role depends on the \textit{locale}'s geometry. We separate two cases.

\textbf{Case 1: the RSU largely covers the scene.}
At Locale~A (RSU occlusion 9\%), all $K_{\mathrm{cav}}{=}4$ selection policies tie on RSU-occluded recall within sampling noise (0.43--0.44, \cref{tab:mech_ablation}). The selector's contribution is purely compute cost reduction: cutting fusion from $K{=}N$ to $K{=}5$ restores feasibility, regardless of which four CAVs are chosen. %

\textbf{Case 2: the RSU has blind spots.}
At Locale~B (RSU occlusion 45\%), the RSU has wide blind spots and CAV selection determines whether occluded objects are recovered. Random selection at $K_{\mathrm{cav}}{=}4$ reaches 0.283 RSU-occluded recall; fusing all $N{-}1$ CAVs reaches 0.349 (a 6.6\% gap that selection must close); the Oracle at $K_{\mathrm{cav}}{=}4$ reaches 0.365 by skipping noisy contributors.

\begin{figure}[t]
  \centering
  \includegraphics[width=\columnwidth]{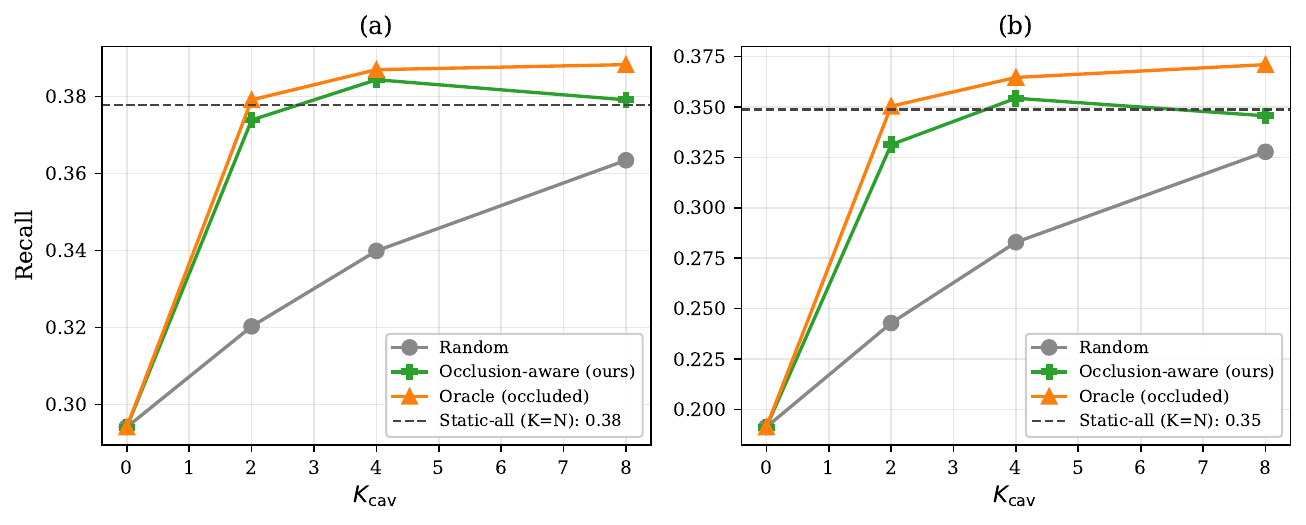}
  \vspace{-6mm}
  \caption{ Total recall (a) and RSU-occluded recall (b) vs.\ $K_{\mathrm{cav}}$ on Locale~B ($\bar{N}{=}12$, RSU occlusion ${\sim}45$\%). %
  {At the high-occlusion \textit{locale}, four well-selected contributors close 87\% of the gap to oracle on RSU-occluded recall.}
  The Occlusion-aware selector tracks the Oracle. %
  }
  \label{fig:selector}
\end{figure}

\textbf{The occlusion-aware selector closes most of the recall gap in Case 2.}
\Cref{fig:selector} shows the Occlusion-aware selector tracking the Oracle on Locale~B. At $K_{\mathrm{cav}}{=}4$, the selector reaches 0.354 RSU-occluded recall. This recovers 87\% of the improvement available between Random selection (0.283) and the Oracle (0.365), while using only four CAVs. The selector slightly exceeds Static-all's recall (0.349) by excluding noisy CAVs. It adds no edge-side forward pass. The selector scores a single BEV-grid tensor reduction over the per-agent classifier maps that fusion already consumes. Simpler score-based top-$K$ selectors (e.g.\ feature-norm) underperformed Random admission in our trials, so we use Random as the deployable lightweight baseline and Oracle as the upper bound.

The full Joint controller uses this same selector plus an adaptive predictor. On the recall axis, Joint and Occlusion-aware produce identical numbers, because the prediction stage does not change detection recall. \cref{fig:selector} therefore omits a separate Joint curve. The Joint controller's superiority lies in AoI compliance (\cref{ssec:joint_controller}) rather than in selector recall. %

\begin{table}[b]
\vspace{2mm} 
\caption{Mechanism ablation for Locale~A ($N{=}23$, 50 updates per config). Envelope compliance (\%): fraction of cycles with composed AoI inside 300\,ms (\cref{ssec:aoi}); mean/p95 refer to the \textit{edge} latency component (not end-to-end latency). Absolute recall is bounded by a model-and-dataset structural ceiling (\cref{sec:architecture}).}
\label{tab:mech_ablation}
\footnotesize
\centering
\setlength{\tabcolsep}{3pt}
\begin{tabular}{l@{\hspace{4pt}}rrrrr}
\toprule
\textbf{Config} & \textbf{mean} & \textbf{p95} & \textbf{env.} & \textbf{R} & \textbf{R$_{\mathrm{occ}}$} \\
                & \textbf{(ms)} & \textbf{(ms)} & \textbf{(\%)} &           &                            \\
\midrule
Static-all ($K{=}N{=}22$)             & 233 & 242 &   0 & 0.590 & 0.447 \\
Random ($K_{\mathrm{cav}}{=}4$)             & 154 & 155 &  98 & 0.587 & 0.434 \\
Occlusion-aware ($K_{\mathrm{cav}}{=}4$)    & 147 & 152 & 100 & 0.592 & 0.438 \\
Joint ($K^*_{\mathrm{cav}}{=}4$, +adapt.)   & 145 & 150 & 100 & 0.592 & 0.438 \\
\bottomrule
\end{tabular}
\end{table}

\subsection{Mechanism Ablation}
\label{ssec:joint_controller}

The ablation separates the two controller effects: selecting fewer CAVs removes the dominant fusion cost, while prediction adaptation provides the remaining compute savings once fusion is feasible.

\Cref{tab:mech_ablation} reports each policy at Locale~A ($N{=}23$), the dense operating point where the cliff is sharpest. Static all ($K_{\mathrm{cav}}{=}22$) costs 233/242\,ms mean/p95, resulting in 0\% envelope compliance; fusing every contributor every cycle is infeasible at this density. Any $K_{\mathrm{cav}}{=}4$ subset reduces edge p95 to ${\leq}155$\,ms and reaches 98--100\% compliance. The compute latency reduction comes entirely from reducing $K_{\mathrm{cav}}$ from 22 to 4. Total recall is essentially unchanged across the $K_{\mathrm{cav}}{=}4$ policies (within sampling noise of Static-all).

Adding the adaptive predictor on top of Filter-only saves only 2\,ms at this operating point (Joint 145/150\,ms vs.\ Filter-only 147/152\,ms), because even after reducing $K_{\mathrm{cav}}$ from 22 to 4, fusion at $\sim$43\,ms is still the largest single edge stage, so prediction-side savings of a few ms have small relative impact. %
The predictor's latency contribution is visible in cross-scale experiments (\cref{fig:cross_scale_compliance}): Prediction-only by itself collapses to 84\% compliance at $N{=}24$, but pairing the same divergence gating and risk-budgeted scheduling with selection allows the Joint controller to hold 100\% across density.

\subsection{Tracking and Prediction Quality}
\label{ssec:trk_pred_quality}

The controller saves prediction compute only if cached and fallback predictions are used on tracks where they do not materially harm forecast quality.

\begin{table}[t]
\centering
\caption{Prediction quality by source under the Joint controller on Multi-V2X. FDE and ADE are computed against ground truth. For MTR-based rows we use the closest of MTR's six predicted trajectories, matching CMP's protocol~\cite{cmp}; linear fallback and stationary reuse emit one trajectory. Moving-track rows use $v{\geq}2$\,m/s; the stationary row uses $v{<}2$\,m/s.
}
\label{tab:pred_quality}
\footnotesize
\setlength{\tabcolsep}{4pt}
\renewcommand{\arraystretch}{0.95}
\begin{tabular}{l@{\hspace{5pt}}rrrrrrr}
\toprule
\textbf{Prediction source} & $n$ & \multicolumn{3}{c}{\textbf{FDE} (m)} & \multicolumn{3}{c}{\textbf{ADE} (m)} \\
\cmidrule(lr){3-5}\cmidrule(lr){6-8}
& & 1\,s & 3\,s & 5\,s & 1\,s & 3\,s & 5\,s \\
\midrule
Fresh MTR        & 552  & 3.48 & 4.19 & 4.64 & 2.61 & 4.04 & 4.64 \\
Cache (amortized) & 495  & 3.67 & 4.32 & 4.66 & 3.04 & 4.28 & 4.83 \\
Linear fallback  & 53   & 2.44 & 8.96 & 8.92 & 1.27 & 4.04 & 6.43 \\
Stationary const.\ & 5,369 & 0.61 & 0.64 & 0.67 & 0.61 & 0.62 & 0.65 \\
\bottomrule
\end{tabular}
\end{table}

Conductor's controller makes adaptive trajectory predictions using different prediction methods, as described in \cref{sec:controller}. 
Re-predicted tracks are derived from fresh MTR computation, while predictions for non-divergent or lower-risk tracks reuse prior cached predictions or are cheaply derived via fallback to linear extrapolation. Objects determined to be stationary ($v{<}2$\,m/s) are predicted to stay immobile.
We evaluate each used method's prediction quality in terms of average and final displacement error (ADE/FDE) compared to ground truth, during a target time horizon. 

\Cref{tab:pred_quality} reports FDE and ADE by prediction method for three time horizons (1, 3, 5 seconds), with $n$ indicating the number of predictions made by each method out of the scenario's total of 6,469 predictions.
Cache reuse tracks fresh MTR closely: FDE is within 0.20\,m at 1\,s and within 0.02\,m at 5\,s.
Linear fallback is used on a small low-risk subset ($<$1\% of predictions); it is very accurate at 1\,s (as it is used on easy-to-predict objects) and drifts to almost 2$\times$ MTR's FDE by 5\,s, due to its constant-velocity extrapolation.
Location prediction for stationary tracks, which dominate the prediction stream (83\%), attains sub-meter accuracy.
These results indicate that the controller's prediction savings do not come from dropping difficult moving tracks. Cache reuse stays close to fresh MTR, linear fallback is rare and used on lower-risk tracks, and stationary tracks remain below 1\,m error across all time horizons.

\subsection{AoI}
\label{ssec:aoi}

\begin{figure}[t]
  \centering
  \vspace{-4mm}
  \includegraphics[width=\columnwidth]{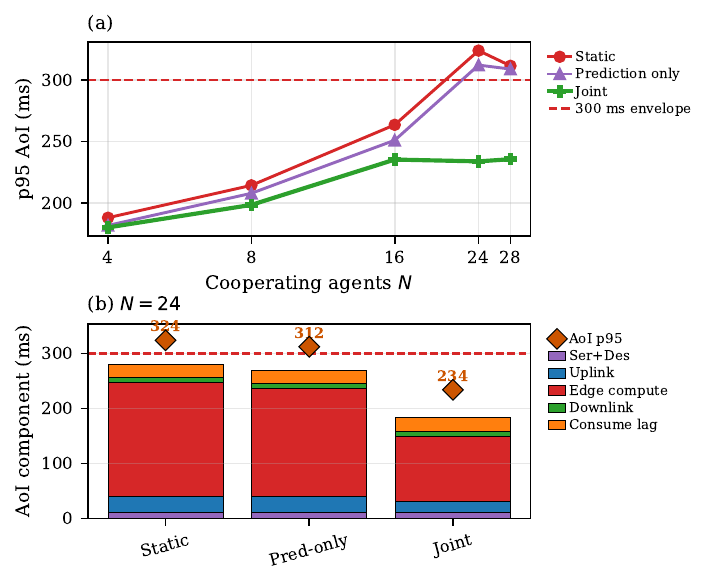}
  \vspace{-8mm}
  \caption{AoI. (a)~p95 AoI vs.\ $N$ per policy; reference lines mark the 300\,ms safety bound and the 130\,ms internal compute SLO. (b)~Mean AoI decomposition at $N{=}24$ stacked by component. Joint stays inside the bound across $N \in [4, 24]$; Static and Prediction-only breach the 300\,ms bound at $N{=}24$. %
  }
  \label{fig:aoi}
\end{figure}

We now compose edge compute with network and planner timing to measure the quantity that matters to the CAV planner: AoI when the world model is consumed.

We define AoI as $t_{\mathrm{consume}} - t_{\mathrm{represent}}$, where $t_{\mathrm{represent}}$ is the timestamp of the scene state the world model reflects. We compose AoI from per-update offline-profiler measurements of edge compute, ns-3 LUT samples for uplink and downlink, and on-A10 calibration for serialize, deserialize, and consume lag (\cref{ssec:network}). This composition preserves compute and network variation instead of adding only mean delays.

\Cref{fig:aoi}(a) reports p95 AoI for each policy and $N{\in}[4,24]$, aggregated across the dense Multi-V2X \textit{locale} set. Static edge p95 rises from 119\,ms at $N{=}4$ to 233\,ms at $N{=}24$, which grows to 311\,ms (i.e., past the 300\,ms bound) after adding the network and consume terms. Prediction-side adaptation reduces MTR cost but leaves fusion uncontrolled, so it tracks the static curve upward and breaches the bound at the same density.

The Joint controller is the only policy that meets the 300 ms deadline. Edge latency stays at 71--123\,ms across $N{\in}[4,24]$ and edge p95 below 150\,ms. Composed AoI p95 ranges from 180\,ms at $N{=}4$ to 234\,ms at $N{=}24$, and envelope compliance is 100\% across the range.

The Joint policy's latency in \Cref{fig:aoi}(a) grows from $N{=}4$ to roughly $N{=}16$ and flattens beyond. The growth is the rolling cost model converging on a steady $K^*_{\mathrm{cav}}$ at the budget boundary. The plateau is the controller hitting that budget: once $K^*_{\mathrm{cav}}$ saturates at the largest feasible value, additional CAVs in the \textit{locale} change neither the active uplink count nor the per-cycle fusion cost. 

\Cref{fig:aoi}(b) decomposes mean AoI at $N{=}24$ into its components. In the Static baseline, edge compute saturates the budget and the composed total breaches 300\,ms even before uplink and downlink latency are added. Under the Joint controller, edge compute remains the largest single latency contributor, but the controller meets the bound by reducing absolute fusion cost rather than rebalancing across stages.
Uplink, downlink, serialize/deserialize, and consume lag each contribute a smaller share, and their sum plus the reduced edge compute latency lands inside the 300\,ms envelope.

\textbf{Network jitter, multicast, and delivery loss.}
The ns-3 Uu samples we draw for uplink and downlink already include per-tick delay variation and PRR, so AoI carries the tail of the joint distribution over edge compute and network delay rather than a sum of means. Sub-50\,ms delivery jitter is absorbed by the planner's 50\,ms consume lag; the residual robustness risk is multicast delivery loss. \Cref{fig:aoi} composes downlink under the multicast model, where Conductor publishes one identical world model per \textit{locale} and downlink delay is independent of recipient count. Per-recipient unicast measured by the same cosim as a counterfactual scales sharply: at $N{=}24$ unicast DL reaches 301\,ms p95 with PRR 0.372, and at $N{=}31$ it reaches 952\,ms with PRR 0.222. The contrast is architectural; a shared world model enables multicast while recipient-specific outputs require a separate transmission per receiving CAV. To isolate loss, we inject independent multicast drops on the composed AoI trace: a dropped publish does not reset AoI, so each miss extends it by 50\,ms (the planner cycle). \Cref{tab:loss_sensitivity} sweeps downlink PRR for the Joint controller. Even at PRR 0.85, p95 AoI increases by only 18\,ms and envelope compliance remains 99.5\%, because the controller retains about 80\,ms of headroom.

\begin{table}[t]
\centering
\caption{Joint controller's sensitivity to multicast delivery loss. Drops extend AoI in 50\,ms planner-cycle increments because the planner reuses the last received world model while operating at 20 Hz.}
\label{tab:loss_sensitivity}
\scriptsize
\setlength{\tabcolsep}{4pt}
\renewcommand{\arraystretch}{0.9}
\begin{tabular}{ccc}
\toprule
\textbf{DL PRR} & \textbf{p95 AoI (ms)} & \textbf{Compliance (\%)} \\
\midrule
1.00 & 220 & 100.0 \\
0.95 & 225 &  99.9 \\
0.90 & 231 &  99.8 \\
0.85 & 238 &  99.5 \\
\bottomrule
\end{tabular}
\end{table}

\subsection{Multicast Sensitivity}
\label{ssec:multicast_stress}

The downlink is one NR-MBS multicast transmission modeled analytically at 9\,ms because the public ns-3 NR module does not implement NR-MBS (\cref{ssec:network}). To test whether the timeliness result hinges on that constant, we recompose every sample under pessimistic 20 and 40\,ms downlink assumptions, covering MBS scheduling and HARQ overheads. Using the unrounded trace, N{=}24 p95 AoI becomes 244 and 264\,ms respectively, with 100\% sample-level compliance across all 188 samples and full-trace compliance of 99.99\%. Unicast remains the counterfactual: per-recipient Uu transmission saturates the base-station scheduler at dense recipient counts.

\subsection{Selection Beyond One Locale}
\label{ssec:multilocale}

The 87\% gap closure of \cref{ssec:contributor_selection} comes from one high-occlusion \textit{locale}. Across 16 evaluated \textit{locales} meeting the participation threshold (median connected CAVs $\geq 8$), the selector closes a median 32\% of the random-to-Oracle occluded-recall gap at $K_{\mathrm{cav}}{=}4$ and 22\% at $K_{\mathrm{cav}}{=}2$ among \textit{locales} with measurable headroom; the \textit{locale} of \cref{fig:selector} sits at the top of the distribution at 88\%. \textit{Locales} where the RSU's elevated view already covers the scene show no recall gap to close, and selection there contributes compute reduction alone. Two \textit{locales} underperform random selection at $K_{\mathrm{cav}}{=}4$; both have low occluded-object counts, making the ratio unstable, and we separate near-zero-denominator \textit{locales} in the reported distribution.

\subsection{Closed-Loop Evaluation}
\label{ssec:closed_loop}

The closed-loop experiments test whether the offline results translate into driving outcomes. They separate three failure modes: Late Fusion (LF) is fast but misses occluded threats; Static feature-level fusion can recover those threats but becomes stale under load; Joint preserves the cooperative perception benefit while keeping AoI inside the bound.

We evaluate along two axes, visibility (occluded vs.\ unobstructed) and load (low vs.\ stressed), and three methods: a late-fusion baseline with YOLOv5 per-agent detection and object-level merge at the edge, the Static cooperative world-model stack (all contributors fused, no runtime control), and the cooperative stack leveraging the Joint controller (same implementation as the one used in \cref{ssec:cliff}--\cref{ssec:aoi}'s offline profiler studies).

\subsubsection{Operational success and the physical safety envelope}
\label{sssec:sop}
A closed-loop run succeeds operationally when the scenario's focal CAV 
(i) completes the scenario without colliding ($S_{\mathrm{coll}}$);  
(ii) does not false-brake ($S_{\mathrm{fp}}$)---i.e., no brake event on a track that has no ground-truth counterpart inside the match radius and time-to-collision (TTC) window; 
and (iii) maintains $\geq60\%$ of target speed ($S_{\mathrm{prog}}$). 
We report per-cell success rate $\Pr[S_{\mathrm{op}}{=}S_{\mathrm{coll}} \wedge S_{\mathrm{fp}} \wedge S_{\mathrm{prog}} {=} 1]$ over several trials on variants of the same scenario, detailed below.

The physical safety envelope $Z$ is the largest AoI for which an RSS-style~\cite{rss} emergency brake still completes before the ego enters the conflict zone. We measure $Z$ empirically by sweeping a single delivery delay on an Oracle baseline that uses ground-truth detections. Operational success holds at $\approx 1.0$ for AoI ${\leq} 280$\,ms, drops to 0.5 near 320\,ms, and reaches 0 above 380\,ms. We adopt $Z{=}300$\,ms as the operative bound, at the inflection point of the sweep.

\subsubsection{Scenario family}
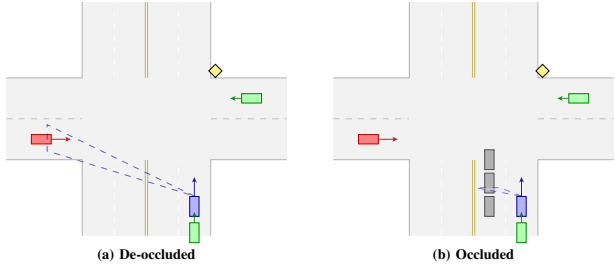
\begin{figure}[t]
  \centering
  \resizebox{\columnwidth}{!}{%
  \begin{tikzpicture}[
      every node/.style={font=\tiny},
      ns car/.style={draw=black, line width=0.25pt, minimum width=1.6mm,
                     minimum height=3.4mm, inner sep=0pt, fill=#1},
      ns car/.default=gray!50,
      ew car/.style={draw=black, line width=0.25pt, minimum width=3.4mm,
                     minimum height=1.6mm, inner sep=0pt, fill=#1},
      ew car/.default=gray!50,
      ego/.style={ns car=blue!30, line width=0.4pt, draw=blue!50!black},
      tesla/.style={ew car=red!50, line width=0.4pt, draw=red!70!black},
      cav ns/.style={ns car=green!30, line width=0.3pt, draw=green!50!black},
      cav ew/.style={ew car=green!30, line width=0.3pt, draw=green!50!black},
      occluder/.style={ns car=gray!60, line width=0.25pt, draw=black!70},
      rsu/.style={draw=black, fill=yellow!60, line width=0.25pt,
                  inner sep=0pt, minimum size=1.6mm},
      sight/.style={draw=#1, dashed, line width=0.35pt, opacity=0.6},
      sight/.default=blue!60!black,
      heading/.style={->, >={Stealth[length=0.8mm,width=0.7mm]},
                      line width=0.3pt, draw=#1, opacity=0.85},
      heading/.default=red!70!black,
    ]

    \def\drawroads{
      \fill[gray!10] (-2.4, -0.7) rectangle (2.4, 0.7);
      \draw[gray!50, dashed, line width=0.2pt] (-2.4, 0) -- (-1.1, 0);
      \draw[gray!50, dashed, line width=0.2pt] (1.1, 0) -- (2.4, 0);
      \fill[gray!10] (-1.1, -2.0) rectangle (1.1, 2.0);
      \draw[yellow!70!black, line width=0.4pt] (-0.02, -2.0) -- (-0.02, -0.7);
      \draw[yellow!70!black, line width=0.4pt] (0.02, -2.0) -- (0.02, -0.7);
      \draw[yellow!70!black, line width=0.4pt] (-0.02, 0.7) -- (-0.02, 2.0);
      \draw[yellow!70!black, line width=0.4pt] (0.02, 0.7) -- (0.02, 2.0);
      \draw[white, dashed, line width=0.3pt] (0.55, -2.0) -- (0.55, -0.7);
      \draw[white, dashed, line width=0.3pt] (0.55, 0.7) -- (0.55, 2.0);
      \draw[white, dashed, line width=0.3pt] (-0.55, -2.0) -- (-0.55, -0.7);
      \draw[white, dashed, line width=0.3pt] (-0.55, 0.7) -- (-0.55, 2.0);
      \fill[gray!10] (-1.1, -0.7) rectangle (1.1, 0.7);
      \draw[gray!60, line width=0.25pt]
        (-2.4, 0.7) -- (-1.1, 0.7) -- (-1.1, 2.0)
        (1.1, 2.0) -- (1.1, 0.7) -- (2.4, 0.7)
        (2.4, -0.7) -- (1.1, -0.7) -- (1.1, -2.0)
        (-1.1, -2.0) -- (-1.1, -0.7) -- (-2.4, -0.7);
    }

    \begin{scope}[xshift=0cm, yshift=0cm]
      \drawroads
      \node[ego] (ego) at (0.82, -1.5) {};
      \draw[heading=blue!60!black] (ego.north) -- ++(0, 0.32);
      \node[tesla] (tesla) at (-1.8, -0.35) {};
      \draw[heading=red!70!black] (tesla.east) -- ++(0.32, 0);
      \node[cav ew] at (1.8, 0.35) {};
      \draw[heading=green!50!black] (1.62, 0.35) -- ++(-0.18, 0);
      \node[cav ns] at (0.82, -1.95) {};
      \draw[heading=green!50!black] (0.82, -1.78) -- ++(0, 0.18);
      \node[rsu, rotate=45] at (1.18, 0.82) {};
      \draw[sight=blue!60!black] (ego.north) -- (-1.7, -0.1) -- (-1.7, -0.55) -- cycle;
      \node[font=\scriptsize\bfseries, anchor=north] at (0, -2.1) {(a) De-occluded};
    \end{scope}

    \begin{scope}[xshift=5.6cm, yshift=0cm]
      \drawroads
      \node[ego] (egoO) at (0.82, -1.5) {};
      \draw[heading=blue!60!black] (egoO.north) -- ++(0, 0.32);
      \node[occluder] at (0.27, -1.5) {};
      \node[occluder] at (0.27, -1.1) {};
      \node[occluder] at (0.27, -0.7) {};
      \node[tesla] (teslaO) at (-1.8, -0.35) {};
      \draw[heading=red!70!black] (teslaO.east) -- ++(0.32, 0);
      \node[cav ew] at (1.8, 0.35) {};
      \draw[heading=green!50!black] (1.62, 0.35) -- ++(-0.18, 0);
      \node[cav ns] at (0.82, -1.95) {};
      \draw[heading=green!50!black] (0.82, -1.78) -- ++(0, 0.18);
      \node[rsu, rotate=45] at (1.18, 0.82) {};
      \draw[sight=blue!60!black] (egoO.north) -- (0.45, -1.18) -- (0.04, -1.18) -- cycle;
      \fill[blue!10, opacity=0.4] (egoO.north) -- (0.45, -1.18) -- (0.04, -1.18) -- cycle;
      \node[font=\scriptsize\bfseries, anchor=north] at (0, -2.1) {(b) Occluded};
    \end{scope}

  \end{tikzpicture}%
  }%
  \vspace{-2mm}
  \caption{LTAP/OD scenario for closed-loop evaluation.  Ego (blue) plans an unprotected left turn against a moving non-AV (red, 13\,m/s, distracted).  RSU (yellow) on NE pole; cooperating CAVs (green) in unrelated lanes.  Visibility axis (a vs.\ b) toggles three occluders on the diagonal sight-line; load axis ($N{=}4$ vs.\ $12$) varies queued CAVs (not depicted, no new threat actors).}
  \label{fig:ltap_scenario}
\end{figure}

We use \cref{{fig:ltap_scenario}}'s LTAP/OD (left-turn across path, opposing direction) scenario from the NHTSA Pre-Crash Typology~\cite{NHTSA_Typology}. The ego CAV approaches the intersection from the south and turns left across the eastbound traffic lanes. Visibility toggles a row of parked vehicles in the northbound left-turn lane that block the ego's diagonal sight line to a non-AV crossing vehicle (eastbound at 13\,m/s). Load varies the number of cooperating CAVs queued at the intersection signal at scenario start ($N{=}4$ low vs.\ $N{=}12$ stressed); these CAVs all uplink to the same edge service, loading its fusion stage. Other parameters (weather, time of day, 50\,km/h target speed, RSU pose) are constant across cells. 

We evaluate 10 variants of the scenario by using different seeds.
Each seed varies the spawn timing of the non-AV crossing vehicle (within a 1\,s window) and the initial poses of the queued CAVs (within their assigned lanes), producing meaningful AoI variation at planner decision time, while preserving the scenario's geometry. 

\begin{figure}[t]
  \centering
  \vspace{-2mm}
  \IfFileExists{floats/fig_closed_loop_matrix.pdf}{%
      \includegraphics[width=\columnwidth]{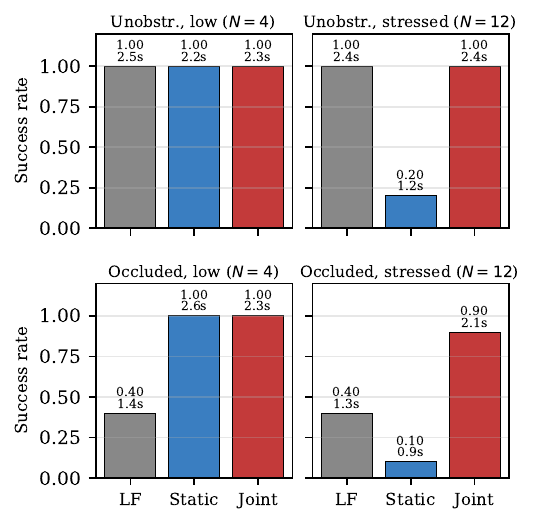}%
  }{%
    \fbox{\parbox{0.9\columnwidth}{\centering\small\textit{[placeholder: closed-loop results pending]}}}%
  }
  \vspace{-7mm}
  \caption{Closed-loop scenario success across the visibility$\times$load matrix. Over-bar numbers: success rate (top) and min-TTC (bottom). %
  }
  \label{fig:closed_loop_matrix}
\end{figure}

\Cref{fig:closed_loop_matrix} demonstrates two distinct claims. \textit{Claim 1: cooperative perception architecture wins where occlusions block late fusion.} Intermediate fusion with a persistent \textit{locale}-level world model recovers RSU-occluded threats that object-level late fusion misses. The occluded low-load cell (bottom left) isolates this effect: LF drops to 0.40 success because no individual agent has a confident detection of the partially-visible threat, while Static and Joint both reach 1.00 by combining partial views in feature space. %
\textit{Claim 2: runtime control preserves that win under high load.} Once the deadline becomes binding, Static loses its perception advantage because the planner needs timely state, while Joint preserves its advantage by keeping the AoI distribution inside the 300\,ms safety bound. The unobstructed-stressed cell (top right) isolates the same effect at low occlusion: Static drops to 0.20 despite direct visibility because the cost of fusion at high $N$ violates the deadline. In the unobstructed-low cell (top left), without occlusion or load stress, all three methods reach 1.00 success. Finally, in the occluded-stressed cell (bottom right) LF fails due to unresolved occlusion (0.40), Static fails due to high$N$ fusion cost (0.10), and Joint is the only method that holds high success rate (0.90). %

The over-bar min-TTC numbers (higher is better) tell the same story from the safety-margin side. Joint keeps min-TTC at 2.1--2.4\,s in every cell, a comfortable braking margin. Static tracks Joint in the low-load cells (2.2--2.6\,s) but drops to $\sim$1.0\,s in both stressed cells, where its success rate also drops to 0.10--0.20. LF's min-TTC drops to 1.3--1.4\,s in both occluded cells because its residual successes happen only when the planner reacts to the occluded threat late. %
In conclusion, Joint is the only configuration that preserves both completion and a healthy reaction margin in every cell. %

\begin{figure}[t]
  \centering
  \IfFileExists{floats/fig_closed_loop_correlation.pdf}{%
    \includegraphics[width=.8\columnwidth]{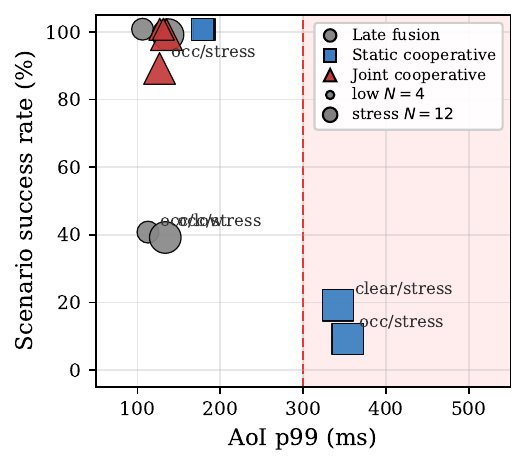}%
  }{%
    \fbox{\parbox{0.9\columnwidth}{\centering\small\textit{[placeholder: closed-loop correlation pending]}}}%
  }
  \vspace{-3mm}
  \caption{Closed-loop success vs.\ AoI, averaged over 10 runs per (visibility, load, method). Marker size: load (small: low $N{=}4$, large: stress $N{=}12$); shape: method. The right-side shaded region violates the $Z{=}300$\,ms deadline. Joint is the only method landing inside the envelope with high success under every condition. %
  }
  \label{fig:closed_loop_correlation}
\end{figure}

\Cref{fig:closed_loop_correlation} binds the offline AoI metric (\cref{fig:aoi}) to the closed-loop outcome. Three patterns are visible.

LF abides by the deadline %
regardless of load: per-agent detection and object-level merge do not scale with $N$. LF's failures (occluded cells, 0.40 success) are architectural, not latency-driven. When no single agent has a confident detection, the merge has nothing to combine. Y-axis spread within LF reflects scenario visibility, not difference in AoI.

Static is bimodal. At low load it meets the deadline and achieves 1.00 success. At stressed load it crosses the deadline at 345 and 351\,ms, and its success collapses to 0.10--0.20 because the world model arrives too stale to act on, %
regardless of visibility.

Joint stays at $\sim$120--150\,ms and $\geq$90\% success in all cases. Its controller adapts $K^*_{\mathrm{cav}}$ to keep the compute budget feasible, so the cooperative-architecture advantage from intermediate fusion is preserved across loads.

\subsection{Additional Conflict Geometries and a Per-Ego Baseline}
\label{ssec:more_scenarios}

We repeat the closed-loop protocol on two further NHTSA conflict geometries and against a directly implemented per-AV arm (\cref{tab:scenarios_ext}). Straight crossing path (SCP): the ego crosses a signalized intersection while a row of parked vehicles occludes crossing traffic. Blind overtake: the ego overtakes a lead vehicle that occludes an oncoming one. Both use the same visibility and load axes, 10 seeds per cell, and each scenario's 300\,ms envelope was independently derived from its conflict geometry and Oracle-verified.

The LTAP/OD pattern reproduces in every geometry. In the occluded stressed cells, late fusion succeeds in 3/10 SCP and 4/10 blind-overtake runs (min-TTCs 1.0 and 1.5\,s), Static in 1/10 of each (0.7 and 1.0\,s), and Joint in 9/10 of each (2.0 and 2.6\,s). In the unobstructed stressed cells Static reaches only 1/10 and 2/10 while Joint reaches 9/10 and 10/10: the deadline failure is independent of visibility.

The per-AV arm replaces Conductor's \textit{locale}-wide execution with CMP-class multi-ego execution: each vehicle fuses on an vehicle-centered canvas and consumes its own output. That canvas sees perpendicular traffic only ${\sim}2.7$\,s before conflict at 14\,m/s, leaving little time for detection, track confirmation, and prediction history. Across 21 occluded LTAP/OD and SCP runs the arm collided in 11, with mean min-TTC 0.5\,s; under the same implementation and conditions Conductor succeeds in 9/10 of each geometry (min-TTCs 2.1 and 2.0\,s). We evaluated the per-AV arm on the two crossing geometries, where canvas extent is the binding constraint.

\begin{table}[t]
\caption{Occluded, stressed cells across conflict geometries: success rate (min-TTC, s) over 10 seeds. CMP-class row: 21 occluded LTAP/OD+SCP runs.}
\label{tab:scenarios_ext}
\footnotesize
\centering
\setlength{\tabcolsep}{4pt}
\begin{tabular}{@{}lccc@{}}
\toprule
\textbf{Method} & \textbf{LTAP/OD} & \textbf{SCP} & \textbf{Blind overtake} \\
\midrule
Late fusion & 0.40 (1.3) & 0.30 (1.0) & 0.40 (1.5) \\
Static cooperative & 0.10 (1.0) & 0.10 (0.7) & 0.10 (1.0) \\
Joint (ours) & \textbf{0.90} (2.1) & \textbf{0.90} (2.0) & \textbf{0.90} (2.6) \\
\midrule
CMP-class per-AV & \multicolumn{2}{c}{0.48 (0.5), 21 runs} & --- \\
\bottomrule
\end{tabular}
\end{table}

\subsection{Failure Modes}
\label{ssec:failure_modes}

Every failure class in the service reduces to one of three planner-visible effects. A \emph{missing threat} arises from detection failure (the model never sees the object) or admission failure (the contributor that saw it was not fused); the occluded closed-loop cells and the selector study measure both. \emph{Stale state} arises from compute or communication delay; the Static stressed cells measure it directly, with world models arriving at 345--351\,ms. A \emph{wrong trajectory} arises from prediction, pose, or timestamp error; divergence gating (\cref{ssec:amortization}) bounds cached-prediction error, and pose or timestamp corruption surfaces as either of the first two effects after fusion. Conductor reduces admission and staleness failures by construction and measures the residual; it does not guarantee collision freedom.

\section{Discussion and Limitations}
\label{sec:discussion}

\textbf{Hardware and deployment scope.}
We evaluate Conductor on an NVIDIA A10 24\,GB edge node. The measured cliff is hardware-specific, but the controller only requires re-measuring the stage-cost coefficients in \cref{eq:cost}. Lower-end accelerators would select fewer CAVs or lower the update rate; larger edge nodes would move the cliff to higher CAV counts. Multi-locale deployments require cross-locale scheduling.

\textbf{Speed-dependent AoI bounds.}
The 300\,ms AoI bound assumes an urban approach speed of 14\,m/s. At 28\,m/s, one lane width of braking reserve corresponds to about 143\,ms. Conductor still applies under a tighter bound, but the controller must select fewer CAVs and run less fresh prediction. Locales whose RSU-only update already exceeds the bound require denser edge placement, faster hardware, or a lower update rate.

\textbf{Evaluation scope.}
Most results come from an offline profiler replaying Multi-V2X point clouds with the same pre-processing, fusion, tracking, prediction, and controller code as the live edge service. These density-scaling results do not close the full detection-to-planner loop. We use the closed-loop platform in \cref{ssec:closed_loop} to validate the main safety effect on representative scenarios. Our profiler spans the Multi-V2X density distribution; behavior beyond this range is not measured.

\textbf{Network robustness.}
The edge uses a bounded jitter buffer to reorder contributor packets before tracking. Our ns-3 Uu samples include delay variation and packet reception ratio, so the AoI results include network variability. The controller uses a fixed 130\,ms compute target and does not yet adapt this margin to recent network delay. A deployment could reserve more compute margin when uplink or downlink delay tails grow, or reduce the fusion set when late packets would make an update stale.

\textbf{Perception quality.}
WorldFusion reaches AP@0.3${=}$0.718, AP@0.5${=}$0.674, and AP@0.7${=}$0.531 on 3{,}504 Multi-V2X car-only frames. On V2X-Sim 2.0~\cite{V2xsim}, the same architecture reaches AP@0.3${=}$0.753, AP@0.5${=}$0.749, and AP@0.7${=}$0.685, comparable to CoBEVT~\cite{xu2022cobevt}. Multi-V2X is harder because it has denser scenes, longer object ranges, and lower per-object point density. Detection misses limit the downstream tracker and predictor.

\section{Conclusion}
\label{sec:conclusion}

Edge-assisted cooperative driving can help CAVs reason beyond line of sight, but only if shared information reaches the planner before it becomes stale. Our work shows that freshness becomes the limiting systems problem when feature fusion, tracking, and trajectory prediction run together at the edge: fusing all available CAVs pushes p95 AoI past the 300\,ms bound beyond 17 participating CAVs. Conductor builds one shared world model for an intersection locale, selects CAVs that add evidence outside the RSU's view, and adapts trajectory prediction work on each update. It keeps AoI within the 300\,ms bound across the density sweep, closes 87\% of the random-to-Oracle gap in the high-occlusion locale, and preserves closed-loop cooperative sensing benefits under both occlusion and dense load.

\section*{Acknowledgments}
This work was funded in part by NSF grant NSF-CNS-2423711 and a gift from Microsoft Corp.

\bibliographystyle{IEEEtran}
\balance
\bibliography{references.bib}

\end{document}